\documentclass[a4paper,fleqn]{cas-sc}

\usepackage[authoryear,longnamesfirst]{natbib}
\usepackage{url}
\usepackage{xcolor}
\usepackage{multirow}
\usepackage{graphicx}
\usepackage{caption}
\usepackage{subcaption}
\usepackage{longtable}
\usepackage{academicons}
\usepackage{threeparttable}
\usepackage{placeins}
\usepackage{float}
\usepackage{amsmath}

\def\tsc#1{\csdef{#1}{\textsc{\lowercase{#1}}\xspace}}
\tsc{WGM}
\tsc{QE}

\begin{document}
\let\WriteBookmarks\relax
\def\floatpagepagefraction{1}
\def\textpagefraction{.001}

\shorttitle{Leveraging Language Prior for Infrared Small Target Detection}

\shortauthors{ P.S. Chib et~al.}

\title [mode = title]{Leveraging Language Prior for Infrared Small Target Detection}

\author[1]{Pranav Singh Chib}[type=editor,
                        auid=000,bioid=1,
                        prefix=,
                        orcid=0000-0003-4930-3937]

\affiliation[1]{organization={Department of Computer Science and Engineering, Indian Institute of Technology Roorkee},
    city={Roorkee},
    state={Uttarakhand},
    country={India}}

\author[1]{Pravendra Singh}[type=editor,
                        auid=000,bioid=1,
                        prefix=,
                        orcid=0000-0003-1001-2219]

\cormark[1]

\ead{pravendra.singh@cs.iitr.ac.in}

\cortext[cor1]{Corresponding author: Pravendra Singh}

\fntext[1]{}


\begin{abstract}
IRSTD (InfraRed Small Target Detection) detects small targets in infrared blurry backgrounds and is essential for various applications. The detection task is challenging due to the small size of the targets and their sparse distribution in infrared small target datasets. Although existing IRSTD methods and datasets have led to significant advancements, they are limited by their reliance solely on the image modality. Recent advances in deep learning and large vision-language models have shown remarkable performance in various visual recognition tasks. In this work, we propose a novel multimodal IRSTD framework that incorporates language priors to guide small target detection. We leverage language-guided attention weights derived from the language prior to enhance the model's ability for IRSTD, presenting a novel approach that combines textual information with image data to improve IRSTD capabilities. Utilizing the state-of-the-art GPT-4 vision model, we generate text descriptions that provide the locations of small targets in infrared images, employing careful prompt engineering to ensure improved accuracy. Due to the absence of multimodal IR datasets, existing IRSTD methods rely solely on image data. To address this shortcoming, we have curated a multimodal infrared dataset that includes both image and text modalities for small target detection, expanding upon the popular IRSTD-1k and NUDT-SIRST datasets. We validate the effectiveness of our approach through extensive experiments and comprehensive ablation studies. The results demonstrate significant improvements over the state-of-the-art method, with relative percentage differences of 9.74\%, 13.02\%, 1.25\%, and 67.87\% in IoU, nIoU, Pd, and Fa on the NUAA-SIRST subset, and 4.41\%, 2.04\%, 2.01\%, and 113.43\% on the IRSTD-1k subset of the LangIR dataset, respectively.
\end{abstract}



\maketitle

\section{Introduction} \label{sec:intro}
InfraRed Small Target Detection (IRSTD) \citep{zhang2022isnet} involves identifying small targets in complex, cluttered backgrounds. The task follows specific criteria \citep{zhang2024irprunedet}: the small target in the infrared image must occupy less than 0.15\% of the total image, the contrast ratio should be below 15\%, and the signal-to-noise ratio should be under 1.5. Detecting small targets in infrared images \citep{dai2021asymmetric,han2020infrared} is challenging due to the small size of the targets, significant infrared energy decay over distances, and the sparse distribution of targets, often resulting in severe imbalances between the objects and background areas.

Traditionally, researchers have used image processing and machine learning techniques to detect small targets, such as filtering techniques, local contrast-based methods, and low-rank based techniques. However, filtering-based methods \citep{dai2017reweighted} can only reduce homogeneous background noise and are unable to mitigate complex background noise. Local contrast-based methods \citep{han2019local} do not perform well when the target is dim. Furthermore, low-rank based methods \citep{rawat2020reweighted} can adapt to low signal-to-clutter ratio images in infrared. However, these methods often have inferior performance \citep{zhang2022isnet}, with high false alarm and miss detection rates, particularly in complex images with complex backgrounds and varied illumination. With the advent of deep learning, infrared small target detection has improved significantly. Detection methods leveraging convolutional neural networks (CNNs) have shown promising performance. Methods such as MDvsFA-cGAN \citep{wang2019miss} aim to reduce false alarms and miss detections using adversarial learning, while ACM \citep{dai2021asymmetric} combines low-level features with high-level semantics. While progress has been made in this field, there is limited exploration of new perspectives on IRSTD using current techniques.

To this end, we explore infrared small target detection from the perspective of multimodal learning that integrates textual information with the imaging modality for the first time to enhance IRSTD capabilities. Integrating information from multiple modalities can significantly enhance the effectiveness of the small target detection task compared to relying on inputs from a single modality. Our approach has the potential to significantly contribute to the development of a multimodal IRSTD network and open up new avenues for future research. 

\begin{figure}[!t]
    \centering
    \includegraphics[scale=.24]{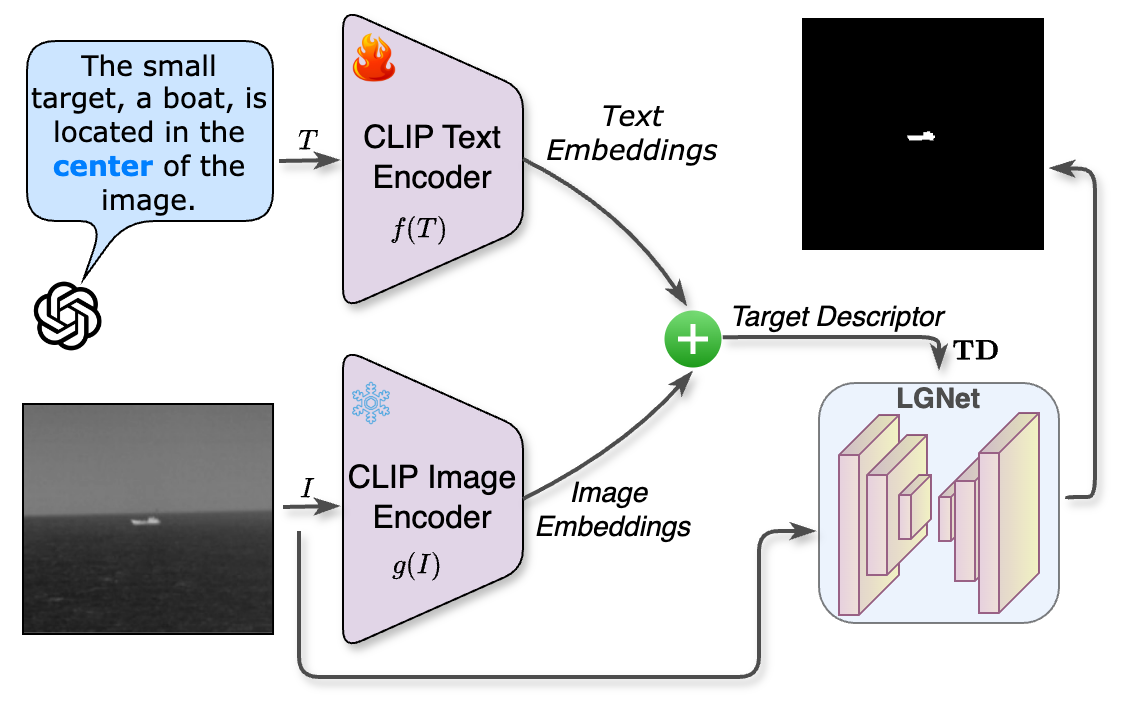}
    
 \caption{Illustration of the language prior used for IRSTD task. The text and image embeddings from CLIP \citep{radford2021learning} are combined element-wise to produce the Target Descriptor, which is then incorporated into our proposed LGNet for small target detection.}
\label{fig:overview} 
\end{figure}

Furthermore, the availability of large vision-language models \citep{lewis2019bart,openai2023gpt,touvron2023llama} and the new learning paradigm of pre-training, fine-tuning, and prediction have shown great effectiveness in various visual recognition tasks \citep{chen2020simple,he2020momentum}. In this paradigm, a pre-trained model is fine-tuned with task-specific annotated training data. Vision-language objectives, such as CLIP \citep{radford2021learning}, are also being utilized in recent models to enable image-text correspondence learning, which improves performance in object detection, semantic segmentation, and related tasks.

Inspired by these advancements, we introduce a novel \textbf{L}anguage \textbf{G}uided \textbf{N}etwork (LGNet) that integrates textual modality with visual data for infrared small target detection (ref. Figs. \ref{fig:overview}, \ref{fig:LGNET}). LGNet enhances IRSTD by incorporating language priors to address the limitations of single-modality approaches by providing high-level semantic information about the target. These priors describe target presence, relative position (e.g., quadrant information), and contextual cues in natural language. When used during training, the language prior guides the model’s attention and improve the learning of more discriminative features in the infrared image.
The language prior is embedded using the CLIP text encoder, while the image embedding is generated by the CLIP image encoder. These text-image embeddings are aligned and then added to form a target descriptor, which is subsequently used in training LGNet for the IRSTD task.

Incorporating the language prior during training significantly improves performance compared to training without it as shown in Table \ref{tab:testtime}. At test time, our method works in both settings: with or without the language prior (see Sec. \ref{sec:target}, \ref{abl:testtime},  and Table \ref{tab:testtime}). This design choice is most suitable considering real-time IRSTD applications, where timely detection is crucial. Therefore, we use the language prior (text description) only during training. During inference at test time, we rely solely on image embeddings as the target descriptor. As shown in Section \ref{abl:testtime}, this approach significantly reduces inference time while maintaining similar performance.

\textbf{Contributions.} Our key contributions are four fold: 
\begin{itemize}
    \item First, we introduce a novel multimodal approach for infrared small target detection that integrates both image and text modalities for the IRSTD task. To the best of our knowledge, this is the first work to propose a multimodal approach for IRSTD. 
    \item Second, we utilize the state-of-the-art vision-language model GPT-4 Vision \citep{openai2023gpt} to generate text descriptions detailing the information about small targets in infrared images, which serves as the language prior. To guide the vision-language model (VLM), we have carefully designed a prompt to ensure accurate detection and precise localization of small targets. 
    \item Third, we comprehensively compare existing infrared small target detection methods and demonstrate that our proposed method significantly enhances small target detection capabilities.
    \item Finally, we have curated a multimodal infrared dataset for the IRSTD task, which includes both image and text modalities.
\end{itemize}

\section{Related Work}
We searched academic databases for related work, including IEEE Xplore, SpringerLink, Elsevier ScienceDirect, and arXiv. The keywords used in our search included: “infrared small target detection,” “IRSTD,” “vision-language models,” “multimodal fusion,” “language-guided detection,” “CLIP,”  “semantic supervision,” and “text-guided object detection.” 

\subsection{Infrared Small Target Detection Methods}

Traditional small target detection methods \citep{zhang2019infrared,han2020infrared,dai2017reweighted,gao2013infrared} use image processing techniques to calculate the infrared image backgrounds and target irregularities. Methods such as the trilayer local contrast measure \citep{zhang2019infrared} and the weighted strengthened local contrast measure \citep{han2020infrared} are commonly used to measure these irregularities. Filtering methods are also employed in traditional systems, including low-rank-based methods like reweighted infrared patch-tensor \citep{dai2017reweighted} and infrared patch-image \citep{gao2013infrared}. Local contrast measure-based methods \citep{chen2013local,han2020infrared} are also used for IRSTD. These methods are ineffective in handling complex targets where the shape and size of the target vary, and the scene is complex with noisy and cluttered backgrounds. Deep neural network methods have emerged to overcome these challenges, learning essential features from infrared images during training. CNNs can understand complex scenes and have significantly improved small target detection performance.  

Some CNN-based methods suffer from the loss of targets in deep layers due to pooling layers. To address this, DNA-Net \citep{li2022dense} was proposed, which uses a dense nested attention network to facilitate gradual interaction between high and low-level features, maintaining small target information in infrared images. In the presence of heavy noise and cluttered backgrounds, targets can be easily lost. ISNet \citep{zhang2022isnet} addresses this issue with the TFD edge block and attention aggregation block, which increase target-background contrast and fuse low-level and high-level information using attention mechanisms bidirectionally. UIU-Net \citep{wu2022uiu} allows multi-level and multi-scale representation learning for objects by incorporating a small U-Net into a larger U-Net backbone. AGPCNet \citep{zhang2023attention} is an attention-guided pyramid context network that fuses contextual information from multiple scales using a context pyramid module, preserving more meaningful information during the upsampling step. DCFR \citep{fan2024diffusion} uses a diffusion-based feature representation network to precisely capture small and low-contrast targets with the help of a conditional denoising diffusion model. RepISD \citep{wu2023repisd} employs different network architectures but maintains equivalent model parameters for training and inference. TCI-Former \citep{chen2024tci} leverages the intrinsic consistency between thermal micro-elements and image pixels, translating heat conduction theories into the network design. Infrared small target detection tends to have larger, intricate models with redundant parameters. To reduce the parameters, IRPruneDet \citep{zhang2024irprunedet} designs a lightweight IRSTD network architecture through network pruning. Deep networks are often considered black boxes, and interpretability is a key issue with such models. RPCANet \citep{wu2024rpcanet} proposes an interpretable network derived from the Robust Principal Component Analysis model, which combines interpretability and data-driven accuracy. RKformer \citep{zhang2022rkformer} is an encoder-decoder structure with a random-connection attention block and Runge-Kutta transformer blocks that are stacked sequentially in the encoder, which enhances features and suppresses noise. Single-frame IRSTD requires pixel-level annotations, which is quite expensive. SSPS \citep{li2023monte} proposes single-point supervision to recover the per-pixel mask of each target using clustering. It is important to learn the shape bias representation for small target detection. SRNet \citep{lin2023learning} and CSENet \citep{lin2024learning} both incorporate shape information into the model learning. Although these methods have shown progress in small target detection, they rely solely on image features, making it difficult to detect small objects in cluttered or noisy backgrounds. We demonstrate that incorporating a language prior alongside the imaging modality significantly improves infrared small target detection performance.

\subsection{Datasets for Infrared Small Target Detection}

Several efforts have been made to develop infrared small target datasets. Wang et al. \citep{wang2019miss} proposed the MFIRST \citep{wang2019miss} infrared dataset, which combines real and synthetically generated images into two subsets: `AllSeqs' and `Single'. The `AllSeqs' subset has image dimensions ranging from 216 × 256 to 640 × 480, while the `Single' subset features images with dimensions between 173 × 98 and 407 × 305. Another popular dataset is NUAA-SIRST \citep{dai2021asymmetric}, which is based on single frames extracted from sequences. It contains 427 images of size 300 × 300, with 55\% of the small targets occupying only 0.02\% of the image area, and approximately 90\% of the images containing a single target. The MFIRST dataset primarily consists of synthetic images, whereas the NUAA-SIRST dataset contains only 427 real images. To address these limitations, the IRSTD-1k \citep{zhang2022isnet} dataset was introduced, featuring 1,001 realistic images of size 512 × 512. Additionally, there are other datasets such as NCHU-SIRST \citep{shi2023infrared}, NUDT-SIRST \citep{li2022dense}, NUST-SIRST \citep{li2022dense}, BIT-SIRST \citep{bao2023improved}, and SIRSTD \citep{tong2024st}.
All the above-mentioned datasets contain only the image modality with the target mask to enable learning models to detect small targets. The recent increase in multimodal approaches, such as text-guided detection \citep{shen2023text,cinbis2012contextual,shangguan2024cross}, has improved the detection capabilities of various models. Shangguan et al. \citep{shangguan2024cross} use textual information to help mitigate domain shift in multi-modal object detection. They utilize a multi-modal feature aggregation module that aligns vision and language for this purpose. Shen et al. \citep{shen2023text} text-guided object detector takes a video frame and text as input, and outputs bounding boxes for objects relevant to the text, improving interpretability and performance across several metrics. Motivated by this, we synthesize a multimodal infrared dataset that includes both image and text modalities for localizing small targets in infrared images. The NUAA-SIRST \citep{dai2021asymmetric} and IRSTD-1k \citep{zhang2022isnet} datasets meet the definition of the Society of Photo-optical Instrumentation Engineers (SPIE) \citep{li2022dense} and are real image datasets, so we built our dataset (LangIR) upon these datasets.

\begin{figure*}[!t]
    \centering
    \includegraphics[scale=.60]{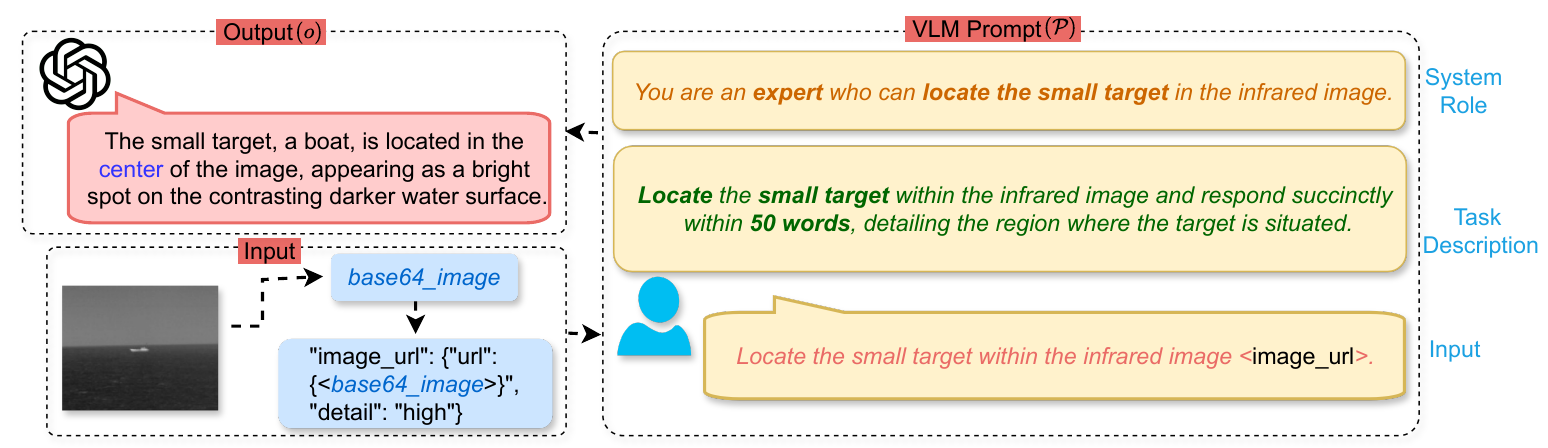}

 \caption{Illustration of the system prompt template utilized for generating textual descriptions using the GPT-4 vision model (VLM). The input image is parsed into a base64 string format and then inserted into the input field of the prompt template. We use VLM in the system role, followed by the task description for the VLM. The prompt is then given to the GPT-4 Vision model to generate a textual description that accurately localizes and describes the small targets in the infrared image.}

\label{fig:main_generation} 
\end{figure*}

\section{Method}

We generate a language prior, which consists of textual descriptions of the target in the infrared image, using a Vision-Language Model (VLM). We employ prompt engineering with varied prompt designs to guide the vision-language model in generating accurate descriptions that localize the target in the infrared image. Existing IRSTD methods, such as MDvsFA \citep{wang2019miss}, ACM \citep{dai2021asymmetric}, DNANet \citep{li2022dense}, RKformer \citep{zhang2022rkformer}, ISNet \citep{zhang2022isnet}, and TCI-Former \citep{chen2024tci}, are designed to work only with image data and cannot incorporate textual descriptions. Therefore, we propose a Language Guided Network (LGNet) that integrates both image and language prior for detecting small targets in infrared images.

\subsection{Data Generation} \label{sec:datagen}
Data generation aims to create descriptive text for a given infrared image. First, we apply the necessary conversions to make these images readable for the VLM (GPT-4 vision model). Each infrared image is encoded into a base64-encoded string, which is then used in the text-based prompt as the final input to the VLM, as shown in Fig. \ref{fig:main_generation}, where the encoded image is passed to the \texttt{image\_url}. 
The resulting textual description of the target is referred to as the language prior in this work. For further details, please refer to the appendix.

\begin{figure*}[!t]
    \centering
    \includegraphics[scale=.12]{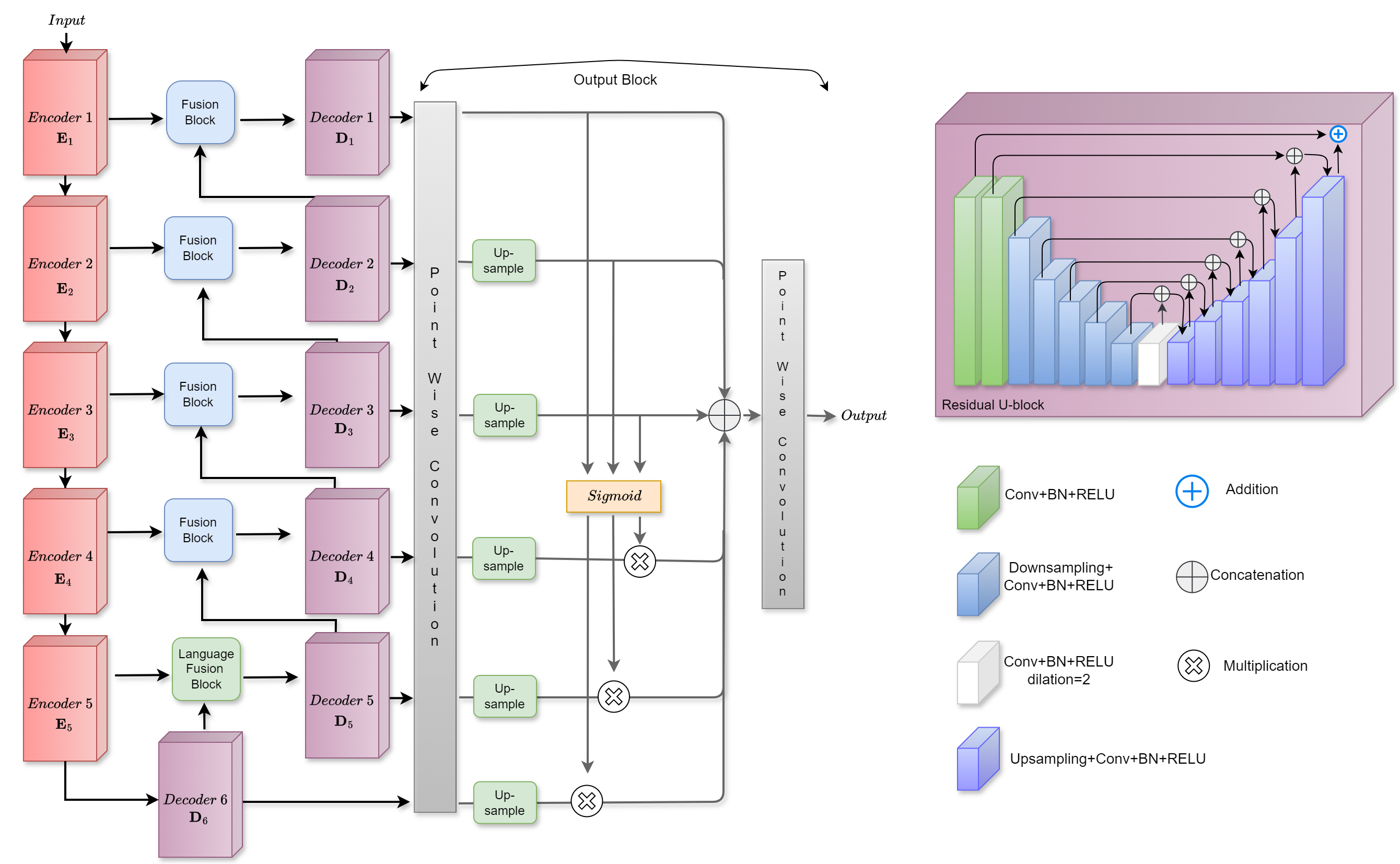}
    \caption{The overview of our proposed LGNet. LGNet is based on UNet architecture with encoder-decoder blocks. Each encoder-decoder is a residual U-Block (shown on the right). The feature maps are encoded gradually, and at the same time, the decoder decodes the feature maps. The outputs of the adjacent encoder-decoders are fused in the fusion block and then passed on to the next decoder. Furthermore, the language prior are used in the language fusion block to get guidance from language.}
  
    \label{fig:LGNET}
\end{figure*}

\subsection{Target Descriptor} \label{sec:target}
The language prior ($T$) is passed as input to the fine-tuned CLIP text encoder \( f(T) \) to obtain text embeddings  $\mathbf{T_e}$, while image embeddings $\mathbf{I_e}$ are generated by the CLIP image encoder \( g(I) \). Building on recent findings \citep{zhou2022extract, zhou2023zegclip} that text embeddings can implicitly align with patch-level image embeddings, we match the text and image embeddings and generate a Target Descriptor $\mathbf{TD}$ by combining them element-wise as \( \mathbf{T_e} + \mathbf{I_e} \). The matching between $T_e$ and $I_e$ is performed by element-wise addition, combining the two embeddings. The resulting Target Descriptor (TD) retains the same tensor size as $T_e$ or $I_e$. During the inference stage, if the language prior is absent, $T_e$ is omitted without affecting the shape of the Target Descriptor, which remains unchanged. This Target Descriptor is then used as input to the LGNet model, enhancing the task of small target detection. The choice of element-wise addition is motivated by its simplicity, efficiency, and compatibility with real-time inference scenarios. In contrast to more complex fusion strategies such as cross-attention, or gated fusion which introduce additional parameters overhead element-wise addition keeps the model lightweight. Importantly, during the inference stage, if the language prior is absent, $\mathbf{T_e}$ can be omitted without affecting the shape of the Target Descriptor. 

\begin{figure*}[!t]
    \centering
    \includegraphics[scale=.15]{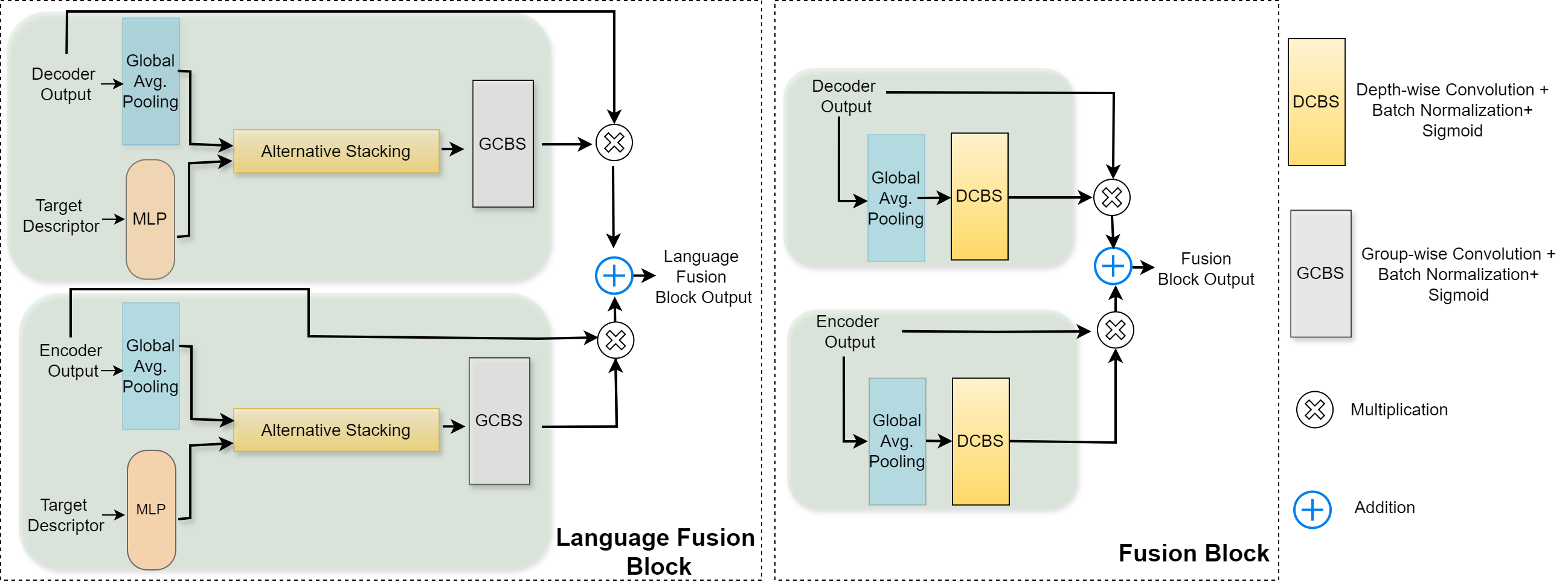}
    \caption{Illustration of the language fusion block (left) and the fusion block (right).} 
  
    \label{fig:FUSIONBLOCK}
\end{figure*}

\begin{figure}[!t]
    \centering
    \includegraphics[width= 0.7\columnwidth]{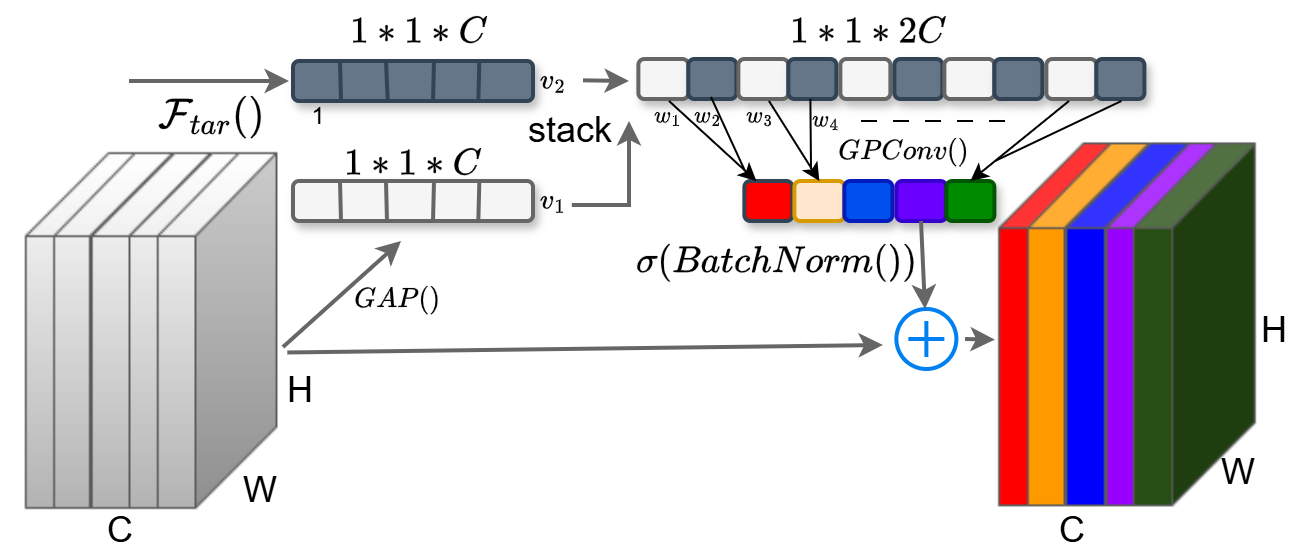}
    \caption{Illustration of the alternative stacking in the language fusion block, where $v_1$ and $v_2$ are combined alternately. This combination then undergoes group-wise convolution, batch normalization, and a sigmoid function to produce the language-guided attention weights.} 
     
\label{fig:alternate stacking} 
\end{figure}

\subsection{LGNet: Language Guided Network}

We propose a Language Guided Network (LGNet) to incorporate both image and target descriptor for the detection of small targets in infrared images. LGNet contains the encoder-decoder structure with residual U-block \citep{qin2020u2} as a structure of the encoder/decoder block. This design facilitates the extraction of both multi-scale and multilevel features. LGNet consists of five encoders $\mathbf{E}_i$ ( $i\in (1,2,3,4,5))$ and six decoders $\mathbf{D}_i$ ( $i\in (1,2,3,4,5,6))$, each of which is a Residual U-block, as shown in Fig. \ref{fig:LGNET}. In residual U-block, the input is first converted into intermediate feature maps. It focuses on extracting local features from the input. The feature maps are gradually downsampled to extract multi-scale features, capturing different levels of detail. These multi-scale features are then progressively upsampled. During this process, the features are concatenated and further processed by convolution to produce high-resolution feature maps. A residual connection is then used intermittently to perform the summation of local and multi-scale features. This summation helps in preserving the details of local features and multi-scale information. Five encoders are used in the encoder stages, and each encoder uses the residual U-block, followed by the downsampling of the feature map. LGNet uses the six decoder blocks. The feature map from the previous decoder is upsampled and fused with the corresponding encoder output in the fusion block (see Fig. \ref{fig:FUSIONBLOCK}), which is then used as input for the next decoder. A fusion block is used to integrate more low-level encoder features into the decoder by generating attention weights. Furthermore, the target descriptor is fused using the language-guided fusion block.

\subsubsection{Language Fusion Block}

To incorporate the target descriptor in LGNet, we propose using a fusion block with language-guided attention. We use language-guided attention only in the last encoder ($\mathbf{E}_5$) and last decoder block ($\mathbf{D}_6$) of the U-shape encoder-decoder structure. As shown in Fig. \ref{fig:FUSIONBLOCK}, the output feature maps from the last encoder are first converted into a vector \( v_1 \) using global average pooling ($\text {GAP}$). Then, the target descriptor is converted into a vector \( v_2 \) of the same size as \( v_1 \) using a Multi-Layer Perceptron ($\mathcal{F}_{tar}$) as shown in Eq. \ref{eq:vector}. 

\begin{equation}
\begin{gathered}
    v_1 =  \text {GAP}(\hat E_5) 
    v_2 = \mathcal{F}_{tar}({\text {TD}})
\end{gathered} \label{eq:vector}
\end{equation}

Finally, the vectors \( v_1 \) and \( v_2 \) are alternately stacked as shown in Fig. \ref{fig:alternate stacking}. This combined vector is processed through group-wise convolution (with group size 2), batch normalization, and a sigmoid function to produce the language-guided attention weights. 

\begin{equation}
\begin{gathered}
\mathbf{F}^{e}_{5} = \sigma (\text{BatchNorm} (GPConv (\text{stack}(v_1, v_2), g=2 ) ) ) \odot \hat E_{\text{5}}
\end{gathered} \label{eq:d1}
\end{equation}

The generated language-guided attention weights are element-wise multiplied with the output feature maps from the last encoder as shown in Eq. \ref{eq:d1}. The same process is applied to the output feature maps from the last decoder (ref. Eq. \ref{eq:d12}), using the language-guided attention weights as shown in Fig. \ref{fig:FUSIONBLOCK}. 

\begin{equation}
\begin{gathered}
\mathbf{F}^{d}_{6} = \sigma (\text{BatchNorm} (GPConv (\text{stack}(v_1, v_2), g=2 ) ) ) \odot \hat D_{\text{6}}
\end{gathered} \label{eq:d12}
\end{equation}

The weighted feature maps from both the encoder ($\mathbf{F}^{e}_{5}$) and decoder ($\mathbf{F}^{d}_{6}$) are added together and then passed to the next decoder as shown in Fig. \ref{fig:LGNET}.

\begin{table}[!t]
    \centering
    \caption{Word counts in LangIR dataset.}
   
    \begin{tabular}{cccccc}
        \toprule
        \textbf{Dataset Subsets} & \textit{Left} & \textit{Right} & \textit{Center} & \textit{Lower} & \textit{Upper} \\
        \midrule
        NUAA-SIRST Train Split & \textcolor{orange}{157} & 87 & 102 & 64 & \textcolor{cyan}{111} \\
        NUAA-SIRST Test Split  & \textcolor{orange}{37}  & \textcolor{cyan}{30} & 24  & 12 & 28  \\
        IRSTD-1k Train Split  & \textcolor{orange}{340} & 193 & 229 & 146 & \textcolor{cyan}{247} \\
        IRSTD-1k Test Split   & \textcolor{orange}{81}  & \textcolor{cyan}{61}  & 53  & 41  & \textcolor{cyan}{61}  \\
        \bottomrule
    \end{tabular}
     
    \label{tab:word_counts}
\end{table}


\subsubsection{Fusion Block} Fusion block also uses the same fusion strategy, where the encoder/decoder output feature maps are converted into a vector using global average pooling followed by depth-wise convolution, batch normalization, and sigmoid function (see Fig. \ref{fig:FUSIONBLOCK}). The generated attention weights are element-wise multiplied with the output feature maps from the encoder/decoder. Then, the weighted feature maps from the encoder and decoder are added together and passed to the next decoder.

\subsubsection{Output Block} As shown in Fig. \ref{fig:LGNET}, the final block is the output block, which fuses the outputs of each decoder. Decoders D4, D5, and encoder D6 outputs are considered as the deep outputs. Decoders D1, D2, and D3 outputs as the shallow outputs. The shallow outputs from the decoder undergo point-wise convolution and are upsampled to obtain the output feature maps from each shallow decoder. Similarly, we obtain the output feature maps from the deep decoders after applying point-wise convolution followed by upsampling. Then, we multiply each feature maps by the scaling weights (of size $\mathcal{R}^{[1,H,W]}$), which we obtain by concatenating the shallow decoder outputs followed by point-wise convolution and applying a sigmoid function (see Fig. \ref{fig:LGNET}). Finally, all the deep and shallow outputs are concatenated, followed by point-wise convolution to get the final output. The final output of the LGNet is the output from the output block.

\subsubsection{Loss Function}
In the training process, we use Binary Cross-Entropy (BCE) loss, which is calculated at each decoder output.
\begin{equation}
    \text{BCE} = {G} \cdot \log(\hat{D_i}) + (1 - G) \cdot \log(1 - \hat{D_i})
\end{equation}

Here, $\hat{D_i}$ represents the output of each $i^{th}$ decoder stage, and $G$ is the ground truth.

\section{LangIR Dataset}

Motivated by the conditional text generation ability of vision-language models with cognitive and reasoning abilities, we construct the \emph{LangIR} dataset for small target detection in infrared images. Our dataset is built upon IRSTD-1k \citep{zhang2022isnet} and NUAA-SIRST \citep{dai2021asymmetric}, two popular real image IR datasets, and we generate textual descriptions that accurately localize and describe the small targets in the infrared images for precise small target detection tasks. The details of the LangIR dataset are given in the following subsection.

\subsection{LangIR Data Statistics}
The proposed dataset contains two subsets, namely NUAA-SIRST and IRSTD-1k. The NUAA-SIRST subset of the LangIR dataset contains 427 text descriptions, while the IRSTD-1k subset contains 1,001 text descriptions. These descriptions include textual information that accurately localizes and describes the small target in the infrared image. In both the NUAA-SIRST and IRSTD-1k subsets, the most frequently occurring positional word is \textit{left}, indicating that most small targets are located in the leftmost position of the image, followed by \textit{upper} and then \textit{center}, as shown in Table \ref{tab:word_counts}.

\section{Experiments}

\subsection{Evaluation Metrics}
We use the same evaluation metrics as prior works \citep{li2023monte,lin2024learning,fan2024diffusion,wu2024rpcanet,zhang2024irprunedet,chen2024tci}. For performance evaluation, we use the following: Intersection over Union (IoU), which calculates the ratio of overlap between the predicted targets and ground truth; It is expressed as given in the following Eq. \ref{eq:iou}: 
\begin{equation}
		\label{eq:iou}
		IoU=\frac{A_i}{A_u}=\frac{\sum\limits_{i=1}^{n}{TP}_i}{\sum\limits_{i=1}^{n}T_i+P_i-{TP}_i},
	\end{equation}
where $A_i$ and $A_u$ are the intersection and union, respectively. $T$ denotes the pixels predicted as the targets. $P$ denotes the pixels of the ground truth targets. $TP$ is the true positive pixels. $n$ represents the number of IR images in the test set.

Normalized Intersection over Union (nIoU), representing the arithmetic mean IoU for each instance in the dataset \cite{dai2021asymmetric}. It is expressed as the following Eq. \ref{eq:niou}:
	\begin{equation}
		\label{eq:niou} 
		nIoU=\frac{1}{n}\sum\limits_{i=1}^{n}\frac{{TP}_i}{T_i+P_i-{TP}_i}.
	\end{equation}

Probability of Detection ($P_d$), which measures the ratio of true targets correctly identified by the model (Eqs. \ref{obj1}); and False-Alarm Rate (Fa), which indicates the proportion of predicted target pixels that do not correspond to any ground truth targets (Eqs. \ref{obj2}).
\begin{equation}
P_d = \frac{1}{n} \sum_{i=0}^{n} \frac{N_i^{\text{pred}}}{N_i^{\text{all}}}
\label{obj1}
\end{equation}

\begin{equation}
F_a = \frac{1}{n} \sum_{i=0}^{n} \frac{P_i^{\text{false}}}{P_i^{\text{all}}}
\label{obj2}
\end{equation}
Here, $N^\text{pred}$ represents the number of correctly detected objects, while $N^\text{all}$ signifies the total number of objects. Similarly, $P^\text{false}$ indicates the pixels of falsely detected objects and  $P^\text{all}$ denotes the total pixels of objects.

\subsection{Implementation Details}

Experiments were carried out using Python 3.8.13 and PyTorch Version 1.13.1 with cu117 support. The training was performed on an NVIDIA RTX A5000 GPU and an AMD EPYC 7543 CPU, with a batch size of 8 across 600 epochs. The Adan optimizer \citep{xie2022adan} was employed, incorporating 10 warm-up epochs and a learning rate of 1e-3. Input images were resized to 512x512 for both training and testing phases. The binary cross-entropy loss function was used with a weight decay of 1e-4. To generate the language prior, we follow the general trend of utilizing the SOTA model GPT-4 Vision\footnote{\url{https://openai.com/api/}}, which is capable of understanding multimodal contexts. However, we have also conducted additional experiments with Claude 3.5 Sonnet\footnote{\url{www.anthropic.com/news/claude-3-5-sonnet}}. Our approach is not restricted to specific models for generating prompts. Instead, the key factor is the use of language prior itself, which significantly enhances IRSTD (Sec \ref{sec:quant}). We used the CLIP ViT-B-16 tokenizer \footnote{\url{huggingface.co/openai/clip-vit-base-patch16}} for encoding both the image and text. The training time of LGNet is 30 seconds per epoch, and the test inference time is 25 milliseconds per infrared image sample. The text generation by the VLM takes around 109 tokens per second\footnote{\url{https://artificialanalysis.ai/models/gpt-4/providers}}. Additionally, the CLIP ViT-B-16 processes each image in 11.6 milliseconds and requires 674 MiB of storage. Our total LGNet parameters are 4.04 million, with FLOPs amounting to 85.68 GFLOPs.

\begin{table}[!t]
\centering
\caption{Ablation study on IRSTD-1k dataset comparing performance across training and testing settings with and without language prior. Never (Baseline): no language prior in training or testing. Training Only: language prior used only during training. Training+Test: language prior used in both training and testing.}
\begin{tabular}{@{}c|c|c|c|c@{}}
\toprule
Availability     & IoU              & nIoU           & Pd             & Fa            \\ \midrule
Never (Baseline) & 69.99            & 67.79          & 94.61          & 6.43          \\ \midrule
Training Only   & 73.30         & 69.09          & 98.26          & 4.09          \\ \midrule
Training+Test    & \textbf{73.35} & \textbf{69.23} & \textbf{98.36} & \textbf{3.89} \\ \bottomrule
\end{tabular}%

\label{tab:testtime}
\end{table}

\begin{figure*}[!t]
    \centering
    \includegraphics[scale=.20]{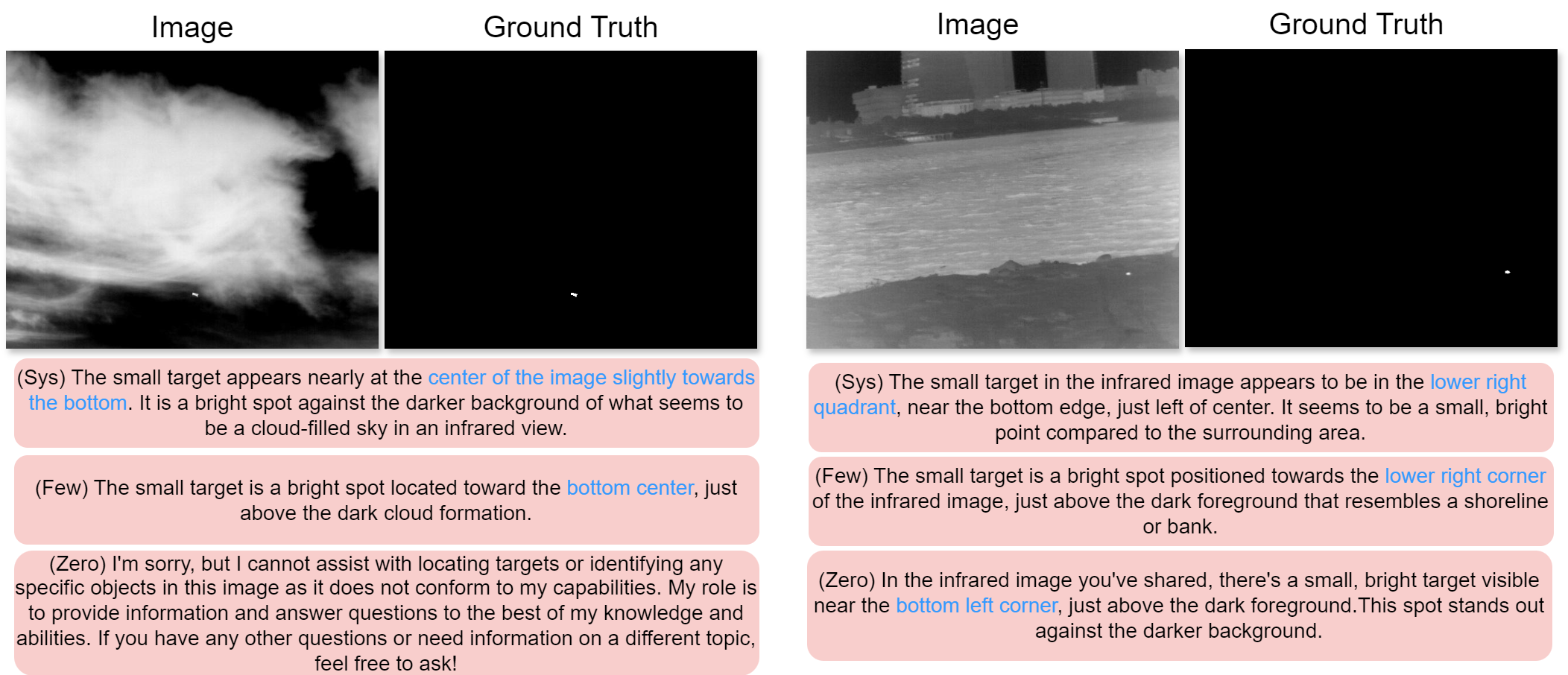}
    
    \caption{Experimental results with three different prompt styles to guide small target detection. `Sys' represents the system prompt, `Few' represents the few-shot prompt, and `Zero' represents the zero-shot prompt. The corresponding ground truth refers to the segmentation mask indicating the actual location of the small target in the infrared image.}
    \label{fig:promptexamples}
   
\end{figure*}

\begin{table}[!t]
\centering
\caption{Performance comparison of various Vision-Language Models (VLMs) on the IRSTD-1K dataset. The table report the Intersection over Union (IoU) and normalized Intersection over Union (nIoU) scores achieved by each model.}
\label{tab:vlms}
\begin{tabular}{c|c|c}
\hline
VLM               & IoU  & nIoU \\ \hline
Our (GPT-4V)      & 73.3 & 69.0 \\ \hline
Claude 3.5 Sonnet & 73.2 & 68.7 \\ \hline
\end{tabular}%
\end{table}

\begin{table}[!t]
\centering
\caption{Comparisons of our approach with State-of-the-Art methods on the IRSTD-1k dataset.}
\setlength{\tabcolsep}{0.35mm}
\begin{tabular}{@{}c|c|c|c|c|c@{}}
\toprule
Methods                               & Venue      & IoU $\uparrow$ & nIoU $\uparrow$ & Pd $\uparrow$ & Fa ($10^{-6}$) $\downarrow$ \\ \midrule
MDvsFA \citep{wang2019miss}            & ICCV19  & 49.50        & 47.41         & 82.11       & 80.33         \\ \midrule
ACM \citep{dai2021asymmetric}          & WACV21  & 58.43        & 54.34         & 89.23       & 23.57         \\ \midrule
DNANet \citep{li2022dense}             & TIP22   & 64.29        & 63.47         & 95.29       & 19.78         \\ \midrule
RKformer \citep{zhang2022rkformer}     & MM22 & 64.12        & 64.18         & 93.27       & 18.65         \\ \midrule
ISNet \citep{zhang2022isnet}           & CVPR22  & 68.77        & 64.84         & 95.56       & 15.39         \\ \midrule
RDIAN \citep{sun2023receptive}         & TGRS23  & 58.85        & -             & 90.48       & 17.76         \\ \midrule
AGPCN \citep{zhang2023attention}       & TAES23  & 57.03        & 52.55         & 88.55       & 16.28         \\ \midrule
RepsISD \citep{wu2023repisd}           & TGRS23  & 65.45        & -             & 91.59       & 7.62          \\ \midrule
UIUNet \citep{wu2022uiu}               & TIP23   & 66.64        & 60.20         & 89.23       & 16.02         \\ \midrule
SRNet \citep{lin2023learning}          & TMM3   & 69.45        & 65.51         & 96.77       & 13.05         \\ \midrule
SSPS \citep{li2023monte}               & ICCV23  & 64.13        & -             & 90.74       & 14.93         \\ \midrule
CSENet \citep{lin2024learning}         & TIP24   & 66.70        & 65.87         & 98.16       & 12.08         \\ \midrule
DCFR \citep{fan2024diffusion}          & TGRS24  & 65.41        & 65.45         & 93.60       & 7.345         \\ \midrule
RPCANet \citep{wu2024rpcanet}          & WACV24  & -            & 63.21         & 88.31       & 4.39          \\ \midrule
IRPruneDet \citep{zhang2024irprunedet} & AAAI24  & 64.54        & 62.71         & 91.74       & 16.04         \\ \midrule
TCI-Former \citep{chen2024tci}         & AAAI24  & 70.14        & 67.69         & 96.31       & 14.81         \\ \midrule
SAIST \citep{zhang2025saist}         & CVPR25  & 72.14        & -         & 96.18       & 4.76         \\ \midrule
IRMamba \citep{zhang2025irmamba}         & AAAI25  & 70.04        & -         & 95.81       & 5.92          \\ \midrule
\begin{tabular}[c]{@{}c@{}} \textbf{LGNet (Ours)}\end{tabular} & -
   &
  \begin{tabular}[c]{@{}c@{}} \textbf{73.30}\end{tabular} &
  \begin{tabular}[c]{@{}c@{}} \textbf{69.09}\end{tabular} &
  \begin{tabular}[c]{@{}c@{}} \textbf{98.26}\end{tabular} &
  \begin{tabular}[c]{@{}c@{}} \textbf{4.09}\end{tabular} \\ \bottomrule
  
\end{tabular}%

\label{tab:irstd}

\end{table}

\begin{table*}[!t]
\centering
\caption{Comparisons of our approach with State-of-the-Art methods on the NUAA-SIRST dataset.}
\setlength{\tabcolsep}{0.5mm}
\resizebox{0.9\textwidth}{!}{%
\begin{tabular}{@{}c|c|c|c|c|c|c|c|c|c|c|c|c|c@{}}
\toprule
Methods &
  \begin{tabular}[c]{@{}c@{}}MDvsFA\\ \end{tabular} &
  \begin{tabular}[c]{@{}c@{}}AGPCN\\ \end{tabular} &
  \begin{tabular}[c]{@{}c@{}}ACM\\ \end{tabular} &
  \begin{tabular}[c]{@{}c@{}}ALCNet\\ \end{tabular} &
  \begin{tabular}[c]{@{}c@{}}UIUNet\\ \end{tabular} &
  \begin{tabular}[c]{@{}c@{}}DNANet\\ \end{tabular} &
  \begin{tabular}[c]{@{}c@{}}DCFR\\ \end{tabular} &
  \begin{tabular}[c]{@{}c@{}}RKformer\\ \end{tabular} &
  \begin{tabular}[c]{@{}c@{}}RDIAN\\ \end{tabular} &
  \begin{tabular}[c]{@{}c@{}}SSPS\\ \end{tabular} &
  \begin{tabular}[c]{@{}c@{}}IRPruneDet\\ \end{tabular} &
  \begin{tabular}[c]{@{}c@{}}$\text{S}^{3}\text{D}$\\ \end{tabular} &
  \begin{tabular}[c]{@{}c@{}}LGNet\\ (Ours)\end{tabular} \\ \midrule
IoU $\uparrow$  & 60.30 & 72.10 & 71.57 & 74.31 & 75.39 & 77.47 & 76.23 &  77.24 & 68.98 & 74.22 & 75.12 & 73.12 & \textbf{82.81} \\ \midrule
nIoU $\uparrow$ & 58.26 & 70.24 & 72.77 & 73.12 & 74.67 & 75.82 & 74.69 &  74.89 & -        & -     & 74.30 & - & \textbf{84.65} \\ \midrule
Pd $\uparrow$   & 89.35 & 80.73 & 98.15 & 97.34 & 97.25 & 98.48 & 99.08 &  99.11 & 96.33 & 96.20 & 98.61 & - & \textbf{99.85}   \\ \midrule
Fa ($10^{-6}$)$ \downarrow$ & 89.35 & 7.23  & 34.47 & 20.21 & 42.41 & 12.86 & 6.520 & 1.58  & 29.63 & 16.19 & 2.96  & 10.28 & \textbf{1.46}     \\ \bottomrule
\end{tabular}%
}

\label{tab:nuaa}
\end{table*}

\begin{figure}[!t]
    \centering
    \includegraphics[scale=.09]{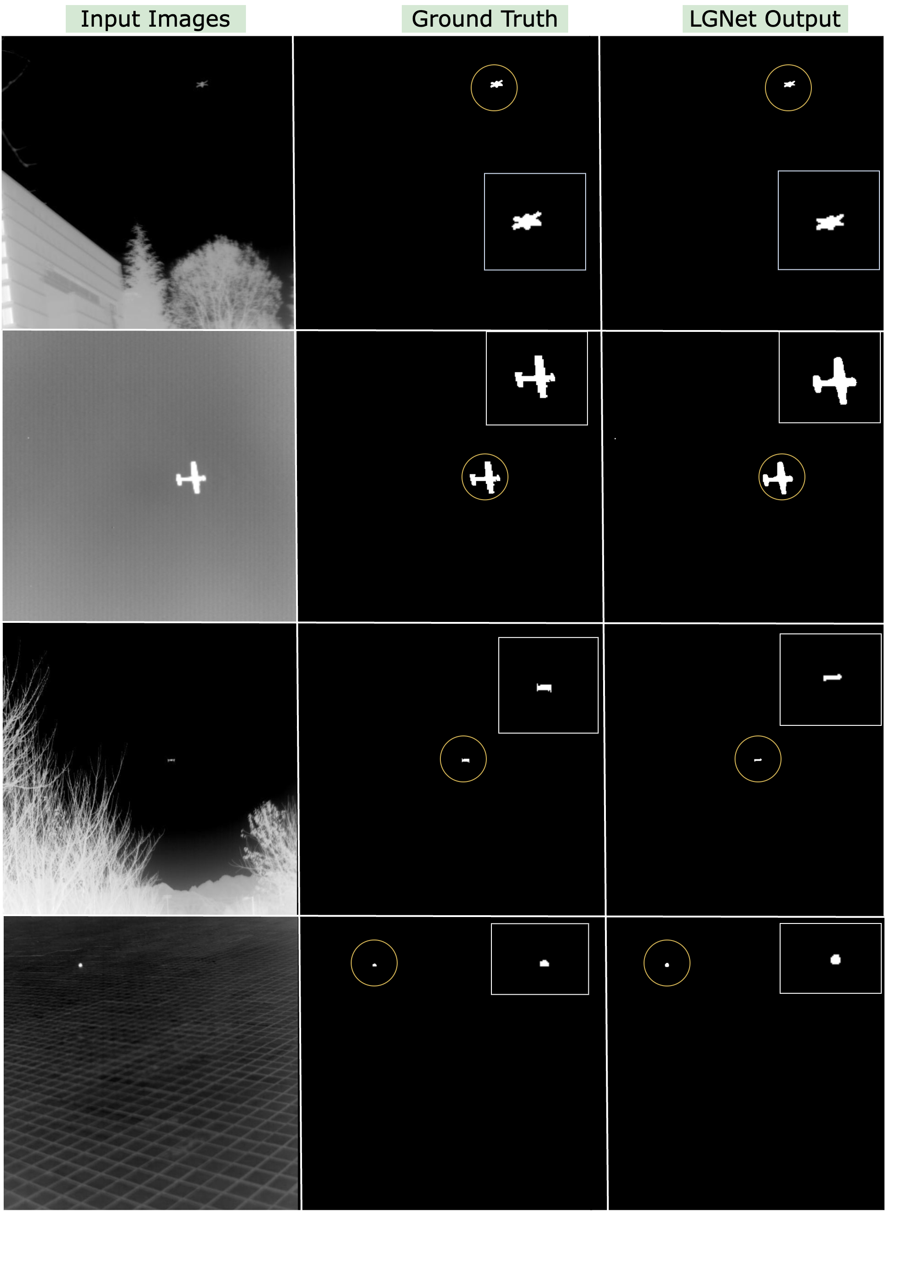}
   
    \caption{Qualitative results obtained using LGNet on the IRSTD-1K dataset. Circles indicate small targets, with an enlarged version shown in the box. The illustration shows LGNet's robustness across varying IRSTD conditions. Row 1 shows detection in a complex background with cluttered trees and structures. Row 2 presents a high-contrast scene. Row 3 depicts a low-contrast target in a natural scene, while Row 4 includes a very small target over a textureless background. LGNet successfully identifies small targets despite background complexity, contrast, and target size variations.}
    \label{fig:lgnet_vis}
  
\end{figure}

\subsection{Experiments on Text Generation}
In this section, we evaluate the text generation component of the proposed framework by assessing the practicality of the textual modality, the effectiveness of different prompt selection strategies.  

\subsubsection{Practicality of textual modality} \label{abl:testtime}

The inference time for LGNet is 25 ms per infrared image, with VLM text generation at around 109 tokens/sec and CLIP ViT-B-16 tokenizing at 11.6 ms per image. In practice, text may be unavailable during test inference, so we apply the language prior only during training. At test time, we rely on image embeddings alone. Table \ref{tab:testtime} shows that even without language priors at test time, our \textit{Training Only} approach performs comparably, supporting real-time IRSTD tasks. LGNet requires ~37 ms (25 + 11.6 ms), while the state-of-the-art TCI-Former \citep{chen2024tci} takes 44 ms, highlighting our model's speed and accuracy in real-world scenarios without language priors. Recent works such as GLIP \citep{li2022grounded}, DETIC \citep{zhou2022detecting}, and TGOD \citep{shen2023text} have demonstrated that incorporating language information during training enhances the learned visual representations, even when language inputs are not available at test inference. Language serves as an auxiliary supervision signal, encouraging the model to learn more discriminative and semantically aligned features. These representations improve model performance, despite the absence of textual input at test inference. Similarly, in our \textit{Training Only} setting, language supervision during training helps the model learn more effective features for small targets in infrared images.

\begin{table}[!t]
\centering
\caption{An ablation study on the IRSTD-1k dataset to show the effectiveness of each component used in our approach.}
\begin{tabular}{@{}c|c|c|c|c@{}}
\toprule
Varient                & IoU   & nIoU  & Pd    & Fa   \\ \midrule
Baseline               & 69.99 & 67.89 & 94.61 & 4.09 \\ \midrule
+Fusion Block          & 70.99 & 67.74 & 94.90 & 3.16 \\ \midrule
+Language Fusion Block & 73.30 & 69.09 & 98.26  & 4.09 \\ \bottomrule
\end{tabular}%

\label{tab:comp}
\end{table}

\subsubsection{Prompt Selection}

 We experimented with three different prompt styles to guide small target detection. Fig. \ref{fig:promptexamples} shows the generated target description texts. The zero-shot chat template contains only the question without any task description or additional role definition. In zero-shot \citep{radford2019language}, we ask GPT-4 Vision to `Detect the small target in the infrared image.' In several instances, the VLM was unable to provide the output and responded with, `As an AI assistant, I am unable to do this task.' To overcome this limitation, we further explore few-shot and system prompts.

In the few-shot prompt \citep{brown2020language}, we provide some examples in the prompt where we describe the small target in the text along with the infrared image. This provides a better task description, and GPT-4 is able to generate more accurate text descriptions. However, there are some instances where GPT-4 generates long descriptions that are irrelevant to the target. Furthermore, the input token size is increased, and we notice a slight increase in the response generation time. To overcome these issues, we use the system prompt  \citep{weston20232} where we define the system role of GPT-4 as an `expert who can locate the small target in the infrared image.' We further refine the response generation by including a well-defined task description: `Locate the small target within the infrared image and respond succinctly within 50 words, detailing the region where the target is situated'.

\subsubsection{Different VLMs and Scalability}
To generate the language prior, we follow the general trend of utilizing the SOTA model (GPT-4V), which is capable of understanding multimodal contexts. We have conducted additional experiments using Claude 3.5 Sonnet for generation. As shown in Table \ref{tab:vlms}, the results are almost similar, indicating that the approach is not restricted to specific models for generating prompts. Instead, the major factor is the use of language prior itself, which significantly enhances IRSTD. This also demonstrates that our framework is model-agnostic and can generalize effectively across various VLMs, including open-source models, enhancing the approach's scalability.

\subsection{Quantitative Results} \label{sec:quant}

We evaluate the performance of LGNet on the IRSTD-1k dataset and compare it with the reported results from previous state-of-the-art (SOTA) methods for infrared small target detection, as shown in Table \ref{tab:irstd}. LGNet achieves significantly better results compared to all other SOTA methods, highlighting the importance of incorporating textual descriptions. Using textual descriptions in addition to image data significantly enhances the performance of target detection models, as evidenced by the achieved IoU of 73.30 and nIoU of 69.09, along with a $P_d$ of 98.26 and $F_a$ of 4.09. Additionally, on the NUAA-SIRST dataset, as shown in Table \ref{tab:nuaa}, LGNet achieves better results compared to other methods such as MDvsFA \citep{wang2019miss}, AGPCN \citep{zhang2023attention}, ACM \citep{dai2021asymmetric}, ALCNet \citep{dai2021attentional}, UIUNet \citep{wu2022uiu}, DNANet \citep{li2022dense}, DCFR \citep{fan2024diffusion}, RKformer \citep{zhang2022rkformer}, RDIAN \citep{sun2023receptive}, SSPS \citep{li2023monte}, IRPruneDet \citep{zhang2024irprunedet}, SAIST \citep{zhang2025saist}, IRMamba \citep{zhang2025irmamba} and $\text{S}^{3}\text{D}$ \citep{zhang2025semi} further underscoring the importance of incorporating textual descriptions.

\subsection{Qualitative Results} 
We visualize the qualitative predictions of LGNet on the IRSTD-1K dataset (see Fig. \ref{fig:lgnet_vis}). The results illustrate that LGNet effectively localizes complex target instances with high precision.
Row 1 shows detection in a complex background with cluttered trees and structures. Row 2 presents a high-contrast scene. Row 3 depicts a low-contrast target in a natural scene, while Row 4 includes a very small target over a textureless background. This demonstrates that LGNet effectively identifies small targets despite variations in background complexity, contrast levels, and target size, even under low signal-to-noise ratios.

\subsection{Ablation Study}


\subsubsection{Impact of Each Components}

We evaluate the effectiveness of each component of LGNet in the task of infrared small target detection. The results are reported in Table \ref{tab:comp}, where we first add the fusion block to the baseline and then apply the language fusion on top of it. In this setting, language was only used during training, not testing, to illustrate a practical scenario where textual input may not be available at inference time. As evident, the language guidance significantly enhances the performance of LGNet. Combining both fusion blocks yields the best results, demonstrating that the language fusion block and the fusion block complement each other. In the reported results, we only use the language prior during training.

\begin{figure}[!t]
    \centering
    \includegraphics[scale=.2]{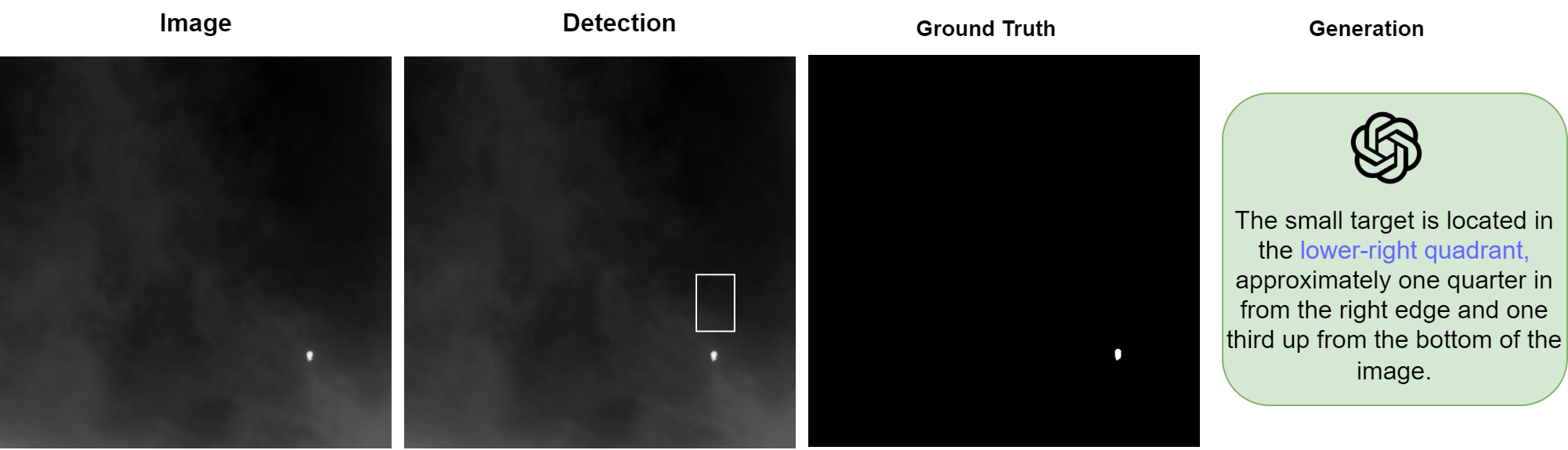}
   
    \caption{Illustration of direct small target detection by the VLM. The VLM is not able to accurately annotate the target because it draws a box that does not include the small target. However, the generated text description is correct, as the VLM uses the term `lower right quadrant' in a broad sense to describe the small target.}
    \label{fig:bounding}
    
\end{figure}

\subsubsection{Directly Detecting Small Target Using VLM in Infrared Image} \label{abl:bounding}
We also experimented with the direct target detection capabilities of the VLM (ref Fig. \ref{fig:bounding}) but did not obtain satisfactory results. The VLM can accurately identify the regions of the target in the form of generated text, such as `small target is located in the center' or `upper left quadrant,' etc. However, when we asked the VLM to exactly annotate the infrared image by creating a bounding box, it failed to do so, as shown in Fig. \ref{fig:bounding}. This is why we chose text descriptions from the VLM instead of direct annotation of the infrared images by the VLM. Furthermore, it should be noted that the VLM can detect the object's location in a broader sense by providing a specific quadrant/region and can draw a bounding box near the small target.

\subsubsection{GroundingDINO + SAM Baseline}

For additional comparison, we finetune a pre-trained GroundingDINO+SAM \citep{ren2024grounded} model for IRSTD. However, the results (on IRSTD-1K dataset), with an IoU of 66.2 and an nIoU of 65.4, are not satisfactory because GroundingDINO+SAM is not designed for small target detection. According to the Society of Photo-Optical Instrumentation Engineers (SPIE), typical infrared small targets have characteristics such as a contrast ratio of less than 15\%, a signal-to-noise ratio of less than 1.5, and a target size of less than 0.15\% of the entire image, which limits the effectiveness of GroundingDINO+SAM in this context.

\subsubsection{Quadrant
Based Generation}

We conducted experiments using only quadrant-based descriptions, where the target’s location in the ground truth determined one of four sentences: The small target lies in the top-left, top-right, bottom-left, or bottom-right quadrant. Without any additional semantic information, the results showed diminished performance (IoU = 72.4, nIoU = 68.1). However, when semantic context was included, performance improved (IoU = 73.30, nIoU = 69.09). This demonstrates that incorporating semantic information enhances the effectiveness of the IRSTD task.


\begin{figure*}[!t]
    \centering
    \includegraphics[scale=.22]{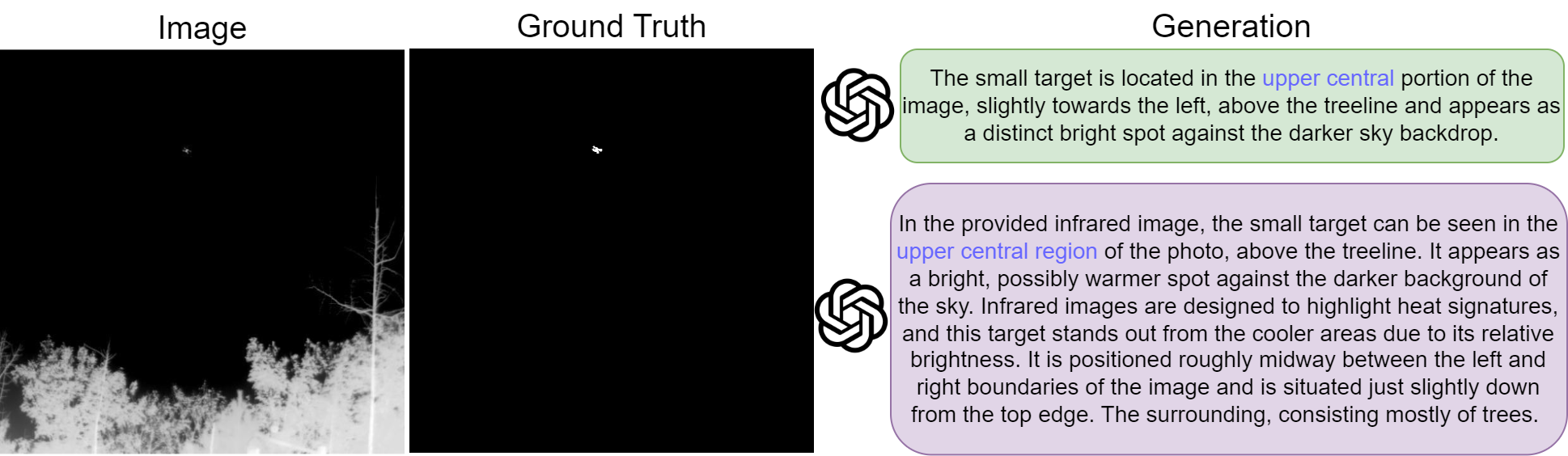}

    \caption{Effect of word limit constraint on responses: the response in \textcolor{green}{green} is generated with a 50-word limit, while the response in \textcolor{purple}{purple} shows the VLM-generated description when no word limit constraint is applied in the prompt.}
    \label{fig:limit}
 
\end{figure*}

\subsubsection{Words Limit Constraint in Response} \label{sec:word_limit}
We also experimented with the length of the generated text (see Fig. \ref{fig:limit}). We compared the generated text by imposing a 50-word limit versus allowing responses with no word limit. Without a word limit, the VLM produces textual descriptions that lack precise information. Therefore, it is important to control the response length to avoid ambiguity and ensure an accurate response from the VLM. It is worth noting that we kept the maximum token size the same and only changed the response limit in the prompt.

\begin{figure*}[!t]
    \centering
    \includegraphics[scale=.16]{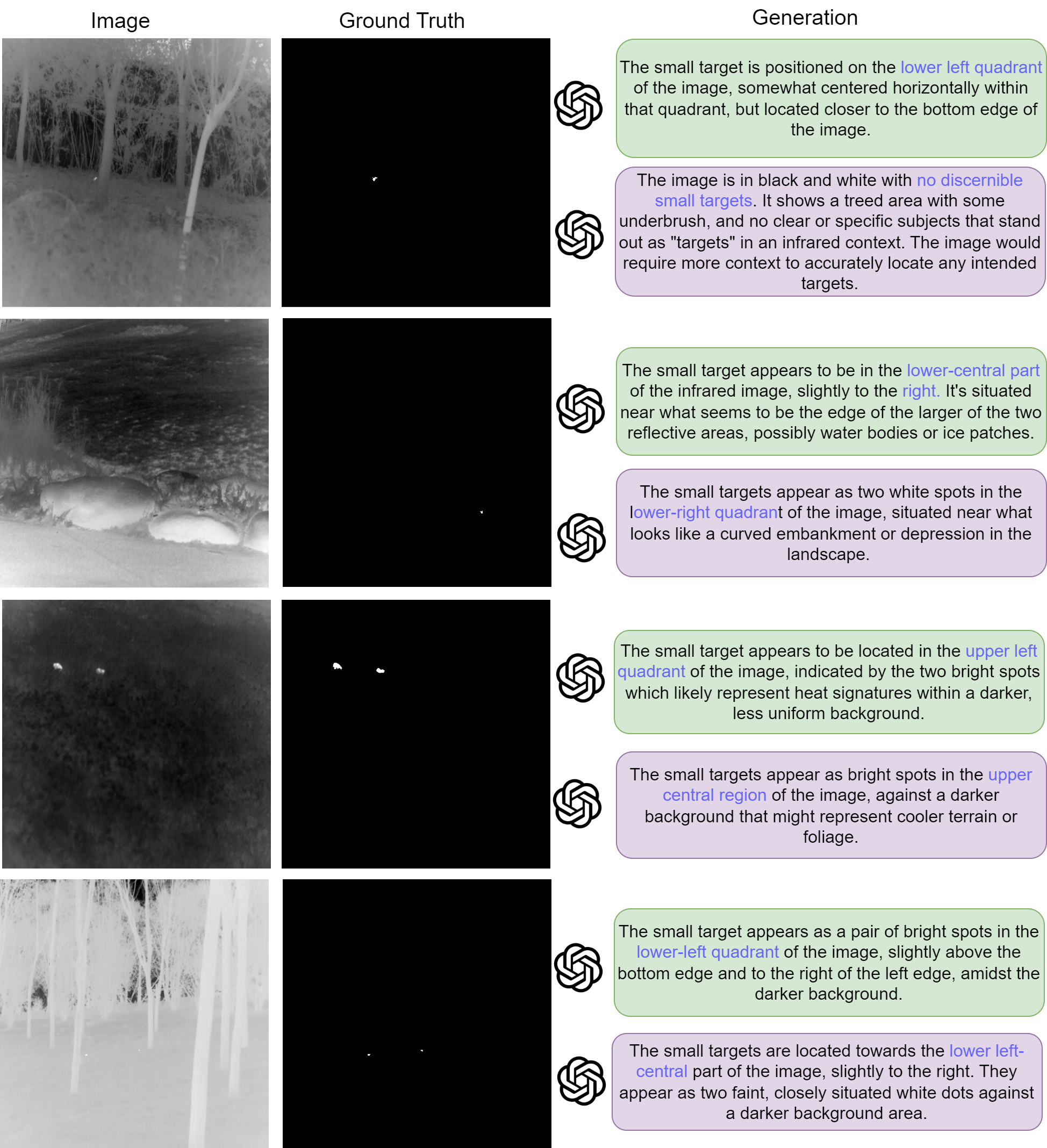}
    \caption{The VLM can detect multiple targets when using the word `target' in the system prompt template, as shown, where it identifies multiple small targets in the response (shown in \textcolor{green}{green color box}) for the IRSTD-1k dataset. The VLM hallucinates when the word `targets' is used in the system prompt template, as shown in the first row (\textcolor{purple}{purple color box}) of the figure. Also, multiple targets are detected by using the word `targets' in the system prompt template, as shown in row 2. Here, the response in the \textcolor{green}{green color box} refers to the generated description when the word `target' is used in the system prompt template, and the response in the \textcolor{purple}{purple color box} refers to the generated description when the word `targets' is used in the system prompt template.}    
    \label{fig:multiple}
\end{figure*}

\subsubsection{Detecting Multiple Small Targets in Infrared Image}

The VLM is capable of detecting multiple small targets in the infrared image (see Fig. \ref{fig:multiple}) when the system prompt template includes the word `target.' However, if the word `target' is changed to `targets' in the system prompt template, the VLM often generates hallucinated descriptions, leading to numerous inconsistent identifications. Therefore, we chose the word `target' instead of `targets' in the system prompt template.

\section{Conclusion}

In this work, we present, for the first time, a novel approach that integrates textual modality with image modality to enhance IRSTD capabilities. Utilizing the state-of-the-art GPT-4 vision model, we synthesize text descriptions detailing the location of small targets in infrared images, employing careful prompt engineering to ensure better accuracy. Language prior is used only during training and not during inference. Since training cost is a one-time overhead, our approach is more scalable in practice at test time. Through extensive experimentation, we demonstrate the effectiveness of our proposed LGNet in incorporating textual descriptions along with image data. We also provide various ablations to validate the efficiency of the VLM in the data generation task.

\clearpage

\printcredits

\bibliographystyle{cas-model2-names}

\bibliography{cas-refs}

\clearpage

\appendix
\section*{Appendix}
\section{Prompt Engineering}

Prompt engineering is the initial step in our data generation process, involving the manual crafting of a suitable prompt. Our well-designed prompt provides the Vision-Language Model (VLM) with a clear understanding of the task to be accomplished, stimulating the model's capabilities to perform infrared small target detection (IRSTD) tasks better.

\section{Prompt Style} 
Our prompt style follows the objective where our designed prompt \(\mathcal{P}\) generates a textual description/annotation ($T$) of the infrared image (i.e., \(T = \mathcal{A}(\mathcal{P})\)). The annotator model (\(\mathcal{A}\)) is the GPT-4 vision model. As shown in Fig. \ref{fig:main_generation}, our prompt style effectively leverages GPT-4 Vision's capabilities to solve the task of identifying small targets in infrared images ($I$). We have included detailed instructions \(\mathcal{I}\) that are well understood by the VLM in the prompt, specifying the system role: ``You are an expert who can locate the small target in the infrared image.'' 

\section{Design Principles} 
We clearly describe the task to avoid any ambiguity and ensure an accurate response from the prompt with word constraints applied to the response. When generating longer responses, the VLM produces textual descriptions containing inaccurate information; therefore, we impose a 50-word limit on the response. Our defined task description is: ``\textit{Locate the small target within the infrared image and respond succinctly within 50 words, detailing the region where the target is situated.}" As experimentally verified, this description accurately conveys the task to the VLM and generates a concise response. The LLM's task is to find the target in the infrared image across diverse scenes with different agents, using heat signatures as a visual cue for the VLM. Furthermore, the word limit constraint ensures that the VLM provides only the most relevant output (ref. 
\ref{sec:word_limit} for ablation on words limit). In this paper, we use the terms ``generated textual descriptions" and ``language prior" interchangeably.

\section{Data Format}

\begin{figure*}[!t]
    \centering
    \includegraphics[scale=.55]{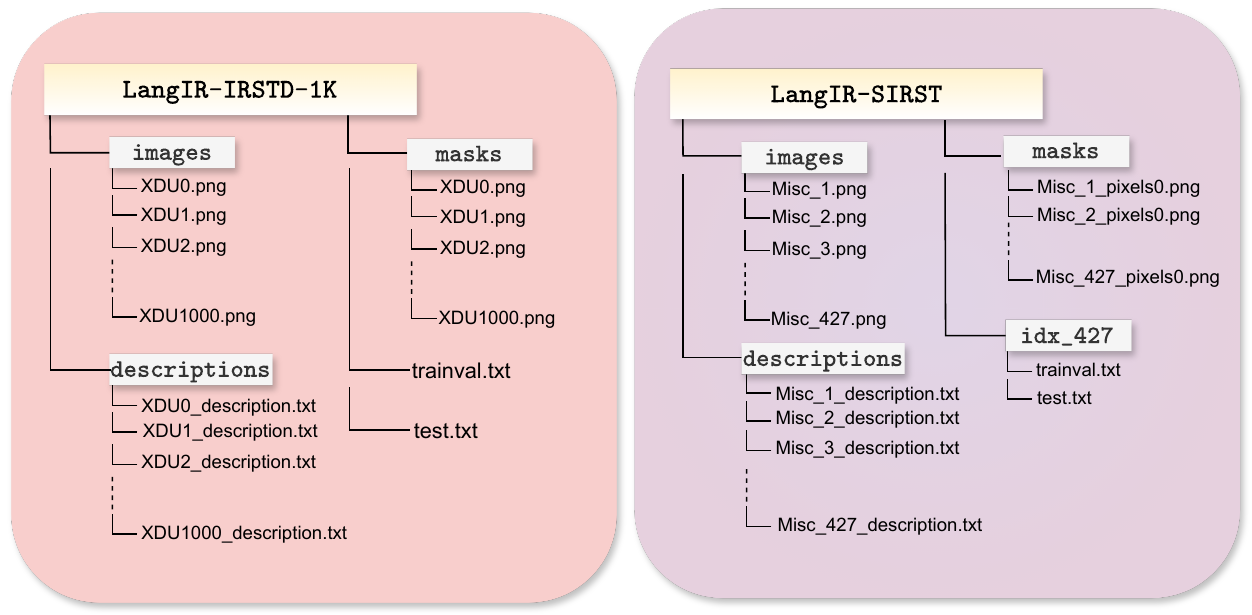}
 \caption{Dataset format of NUAA-SIRST and IRSTD-1k subsets of LangIR. IR images are in the \texttt{images} directory, with the corresponding detection masks in the \texttt{masks} directory. The text descriptions of the small targets are in the \texttt{description} directory.}
 \label{fig:dadasetformat}
\end{figure*}

We follow the same data format as used by prior works \citep{zhang2022isnet,dai2021asymmetric}. LangIR contains two subsets: \texttt{LangIR\_IRSTD} and \texttt{LangIR\_SIRST}, as shown in Fig. \ref{fig:dadasetformat}. In \texttt{LangIR\_IRSTD-1k}, \texttt{images} directory contains the infrared images named XDU$<$\textit{id}$>$.png, where \textit{id} defines the image identifier, starting from 0 and spanning up to 1000. The textual descriptions for small target detection are stored in the \texttt{descriptions} directory, named XDU$<$\textit{id}$>$\_description.txt. The ground truth masks are stored in the \texttt{masks} directory, following the same naming convention as the \texttt{images} directory. The trainval.txt and test.txt files contain the \textit{id}s used for training or testing. 

The \texttt{LangIR\_SIRST} subset of  LangIR contains a total of 427 infrared images in the \texttt{images} directory, each named Misc\_$<$\textit{id}$>$.png. The masks are stored in the \texttt{masks} directory, named Misc\_$<$\textit{id}$>$\_pixels0.png. The textual descriptions for small target detection are stored in the \texttt{descriptions} directory, named Misc\_$<$\textit{id}$>$\_description.txt. The trainval.txt and test.txt files contain the \textit{id}s used for training and testing.

\section{Detecting small targets using VLM}
\begin{figure*}[!t]
    \centering
    \includegraphics[scale=.14]{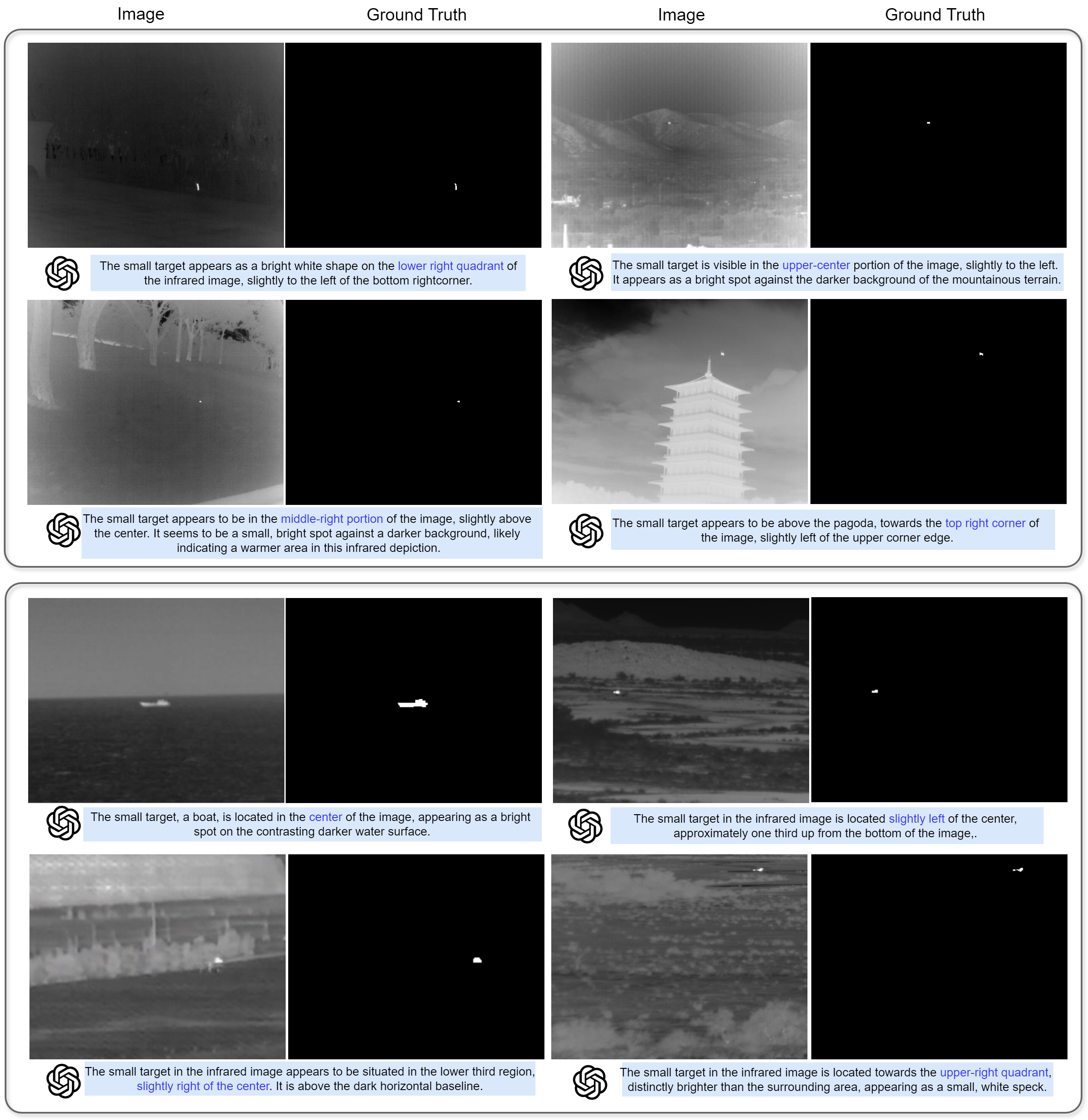}
 \caption{Illustration of the textual descriptions for small target detection in infrared images. (Top) The textual description of the targets in the IRSTD-1k dataset indicates the target object's location in the text. (Bottom) The textual description of the targets in the NUAA-SIRST dataset, where the target location is described.}
 \label{fig:gptresponses}
\end{figure*}

We also attempted to use the GPT-4 vision model to directly detect and create bounding boxes around small targets in infrared images (ref. Sec. \ref{abl:bounding}). However, the VLM was unable to accurately perform this task, and the annotations were incorrect. On the other hand, textual descriptions for small target localization were accurately generated by the VLM because the words used in the response cover target locations in a broader sense, as illustrated in Figs \ref{fig:gptresponses}.

\end{document}